\documentclass{article}
\usepackage{amsmath,epsfig}
\usepackage[preprint]{spconfa4}
\usepackage{xcolor}
\usepackage{cite}

\usepackage{booktabs}
\usepackage{cite}
\usepackage{makecell}
\usepackage{color}

\usepackage[colorlinks,
            linkcolor=black, 
            anchorcolor=black,
            citecolor=green,
            ]{hyperref}

\let\OLDthebibliography\thebibliography
\renewcommand\thebibliography[1]{
  \OLDthebibliography{#1}
  \setlength{\parskip}{0pt}
  \setlength{\itemsep}{0pt plus 0.3ex}
}

\begin{document}\sloppy

% Example definitions.
% --------------------
\def\x{{\mathbf x}}
\def\L{{\cal L}}

% Title.
% ------
\title{PAMI-AD: AN ACTIVITY DETECTOR EXPLOITING PART-ATTENTION AND MOTION INFORMATION IN SURVEILLANCE VIDEOS}
%
% Address.
% ---------------
\name{Yunhao Du$^{1}$, Zhihang Tong$^{1}$, Junfeng Wan$^{1}$, Binyu Zhang$^{1}$ and Yanyun Zhao$^{1,2}$}
\address{
  $^1$Beijing University of Posts and Telecommunications \\
  $^2$Beijing Key Laboratory of Network System and Network Culture, China \\
  {\tt\small \{dyh\_bupt,tongzh,wanjunfeng,zhangbinyu,zyy\}@bupt.edu.cn}
}

\maketitle

\begin{abstract}
  Activity detection in surveillance videos is a challenging task caused by small objects, complex activity categories, its untrimmed nature, etc.
  Existing methods are generally limited in performance due to inaccurate proposals, poor classifiers or inadequate post-processing method.
  In this work, we propose a comprehensive and effective activity detection system in untrimmed surveillance videos for person-centered and vehicle-centered activities.
  It consists of four modules, i.e., object localizer, proposal filter, activity classifier and activity refiner.
  For person-centered activities, a novel part-attention mechanism is proposed to explore detailed features in different body parts.
  As for vehicle-centered activities, we propose a localization masking method to jointly encode motion and foreground attention features.
  We conduct experiments on the large-scale activity detection datasets VIRAT, and achieve the best results for both groups of activities.
  Furthermore, our team won the 1st place in the TRECVID 2021 ActEV challenge.
\end{abstract}
\begin{keywords}
  Activity detection, surveillance videos, attention mechanism
\end{keywords}

\begin{figure}[t]
  \centering
  \includegraphics[width = 0.35\textwidth]{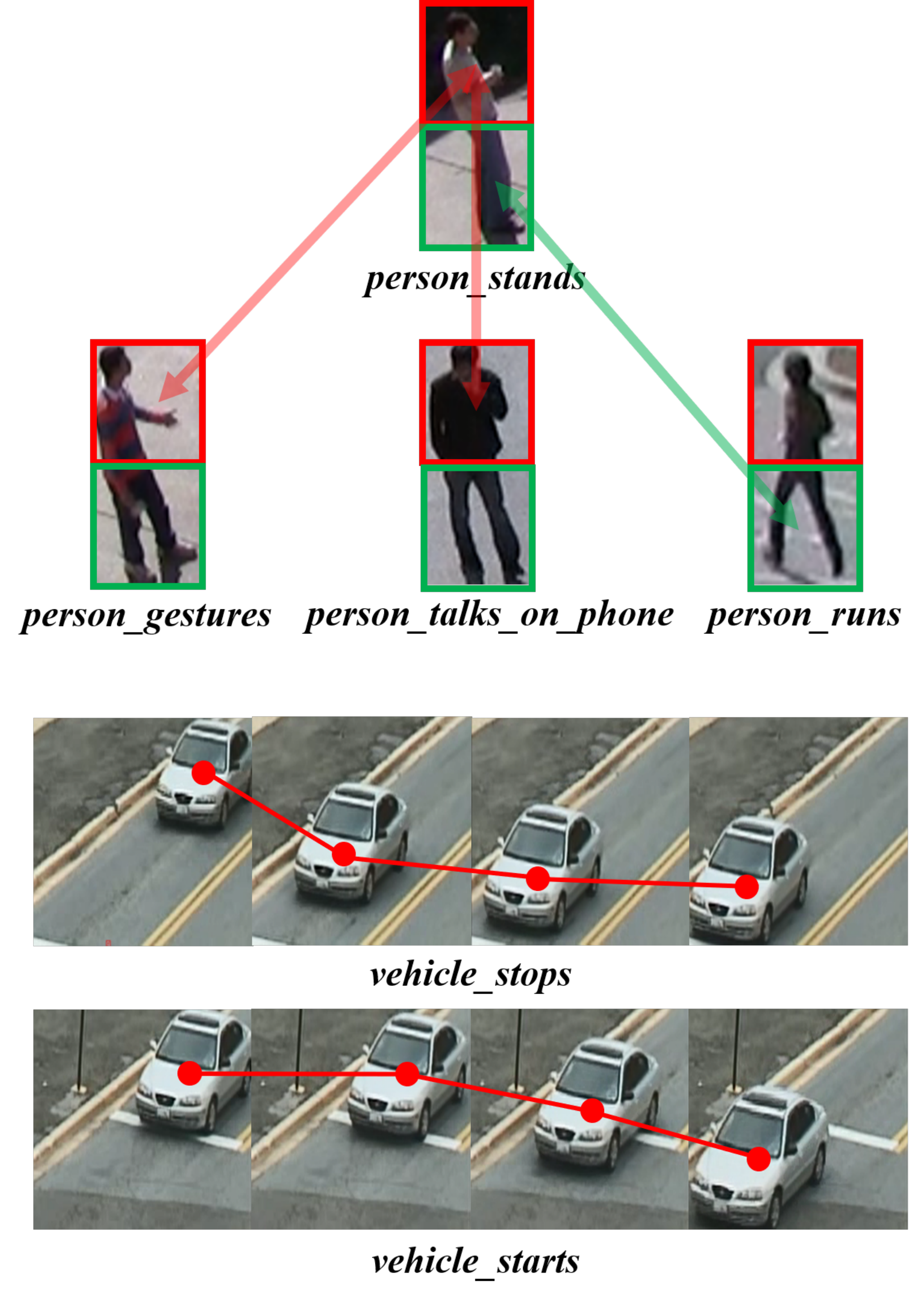}
  \caption{
    Motivations of our method.
    1) The top half shows examples of person-centered activities.
    Different categories can be distinguished with detailed information in different body parts.
    2) The bottom half illustrates the importance of motion information for vehicle-centered activity classification.
    Red dots represent the center of the vehicle.
  }
  \label{figure_motivation}
  \end{figure}

\section{Introduction}
\label{sec:intro}
  In recent years, impressive advancements have been made for action recognition and detection in trimmed videos.
  However, some challenges make it difficult to transfer these works to activity detection in untrimmed videos, 
  like surveillance videos (e.g., the VIRAT dataset \cite{oh2011large}).
  First, we must localize the activity regions spatio-temporally. 
  Second, actors tend to be small, which brings the risk of false detection and missing detection. 
  Third, the diversity and complexity of activity categories requires us to design targeted strategies.
  
  To tackle the above issues, we propose a comprehensive activity detection system in untrimmed surveillance videos,
  especially for person-centered and vehicle-centered activities.
  It consists of four modules, i.e., object localizer, proposal filter, activity classifier and activity refiner.
  First, all objects are localized and tracked by multiple-object trackers \cite{wojke2017simple, bochinski2017high} following the tracking-by-detection paradigm.
  Then, these trajectories are used to generate spatio-temporal proposals.
  The proposal filter module is applied to filter out static objects, which helps reduce the number of false positive proposals.
  Third, two individual classifiers are utilized to classifiy person-centered and vehicle-centered activities separately.
  Finally, we introduce a temporal-NMS based activity refiner to obtain more accurate results.

  As for activity classifiers, existing methods generally apply state-of-the-art models 
  in the action recognition task \cite{tran2018closer, zhou2018temporal} without any specific optimization.
  Differently, we elaborately design the model for the two groups of activities.
  Figure \ref{figure_motivation} illustrates our motivations.
  For person-centered activities, there are only minor differences between different categories.
  Thus we take it as a fine-grained classification task and try to explore detailed information.
  We observe that each body part should have different importance.
  As shown in the top half in Figure \ref{figure_motivation}, "person\_stands" shares similar lower body features 
  with "person\_gestures" and "person\_talks\_on\_phone", but they have different upper body features.
  Instead, we can easily distinguish "person\_stands" and "person\_runs" by lower body features.
  Therefore, we propose a novel part-attention mechanism to explore detailed features in each body part and assign them importance weights for different classes.

  For vehicle-centered activities, motion information palys a more crucial role than details.
  The bottom half in Fugure \ref{figure_motivation} shows the examples of "vehicle\_stops" and "vehicle\_starts" activities,
  where red dots represent the center of the vehicle.
  They share similar appearance features and background information, but we can distinguish them with the help of position change patterns.
  Motivated by this, we design a simple but effective localization masking method which jointly encodes motion and foreground attention features.
  Note that both our part-attention and localization masking method bring negligible computational cost and can be plugged into various models.

  We evaluate our method on the VIRAT dataset, which is a large-scale realistic and challenging surveillance video dataset for activity detection. 
  Experiments show that our method outperforms all the previous works for both person-centered and vehicle-centered activities, 
  which demonstrates its effectiveness.
  Specifically, our proposed part-attention and localization masking strategies outperform the baseline by a large margin.

  The contribution of this paper can be summarized as follows:

  1) We propose a comprehensive system for activity detection in untrimmed surveillance videos.
  
  2) We propose a novel part-attention mechanism for person-centered activities, which makes the model be capable of capturing detailed features of each body part.

  3) We propose a simple but effective localization masking method for vehicle-centered activities, which encodes motion and foreground information simutaneously.

  4) Our system achieves the best results on the VIRAT dataset and our team wins the 1st place in the TRECVID 2021 ActEV challenge.

\section{Related Work}
  \textbf{Action Recognition.}
  Given a video clip, action recognition aims to assign it one or multiple action labels. 
  Many methods are based on two-stream model \cite{simonyan2014two}, 
  which processes the RGB and optical flow separately and fuses them in upper layers \cite{christoph2016spatiotemporal,feichtenhofer2017spatiotemporal}. 
  Others use only one stream to extract features and perform classification, 
  which typically apply 2D-CNN \cite{lin2019tsm}, 3D-CNN \cite{carreira2017quo} or Transformer \cite{fan2021multiscale}. 
  We use MViT \cite{fan2021multiscale} as the baseline to extract spatio-temporal features of activity proposals, due to its powerful encoding ability.

  \noindent \textbf{Action Detection.}
  Different from action recognition, action detection asks for both action labels and temporal/spatial boundaries.
  Many methods take advantage of frame-level detectors (e.g., Faster R-CNN \cite{ren2015faster} and SSD \cite{liu2016ssd}), 
  and propose end-to-end action detection frameworks \cite{xu2017r,lin2017single}. 
  Another intuitive approach is to refine the action boundaries \cite{shou2017cdc,zhao2017temporal}. 
  Some recent works focus on extracting finer features temporally or spatially \cite{piergiovanni2018learning,piergiovanni2019temporal}. 
  However, they are all designed for high-quality videos, not suitable for complex scenarios like surveillance videos.

  \noindent \textbf{Activity Detection in surveillance videos.}
  Gleason et al. perform activity detection based on spatio-temporal proposals and optical flow \cite{gleason2019proposal}.
  Then they improve upon it to achieve real-time performance \cite{gleason2020real}. 
  Rizve et al. propose an encoder-decoder based localization network to detect proposals, 
  followed by a classification network and a post-processing module \cite{rizve2021gabriella}. 
  Yu et al. develop a dense spatio-temporal proposal generation model for more accurate boundaries and win the 1st place in the TRECVID 2020 ActEV challenge \cite{yucmu}.
  However, all these works neglect to design specific strategies for different activity categories. 
  In contrast, we propose a novel part-attention mechanism for person-centered activities and an effective motion information encoding method for vehicle-centered activities.

\begin{figure*}[t]
  \centering
  \includegraphics[width = 0.85\textwidth]{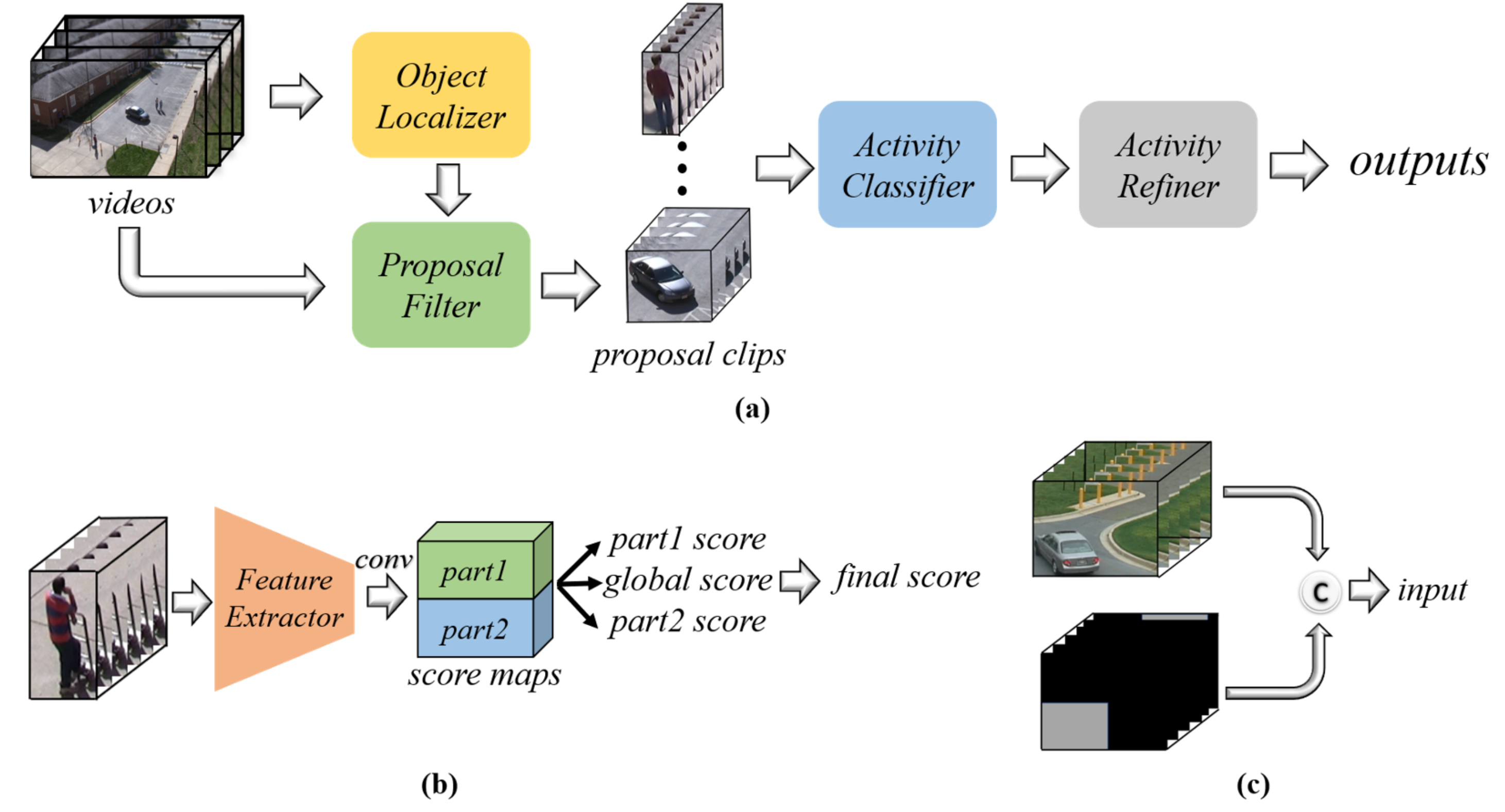}
  \caption{
    Overview of the proposed activity detection framework and optimization strategies. 
    (a) shows our framework which consists of four modules, i.e. object localizer, proposal filter, activity classifier and activity refiner.
    (b) illustrates the perosn-centered activity classifier with part-attention mechanism.
    Note that different scores are aggregated into the final score with learnable weights for different categories.
    (c) illustrates our localization masking method to generate the input of the vehicle-centered activity classifier.
  }
  \label{figure_framework}
  \end{figure*}

\begin{figure*}[t]
  \centering
  \includegraphics[width = 0.95\textwidth]{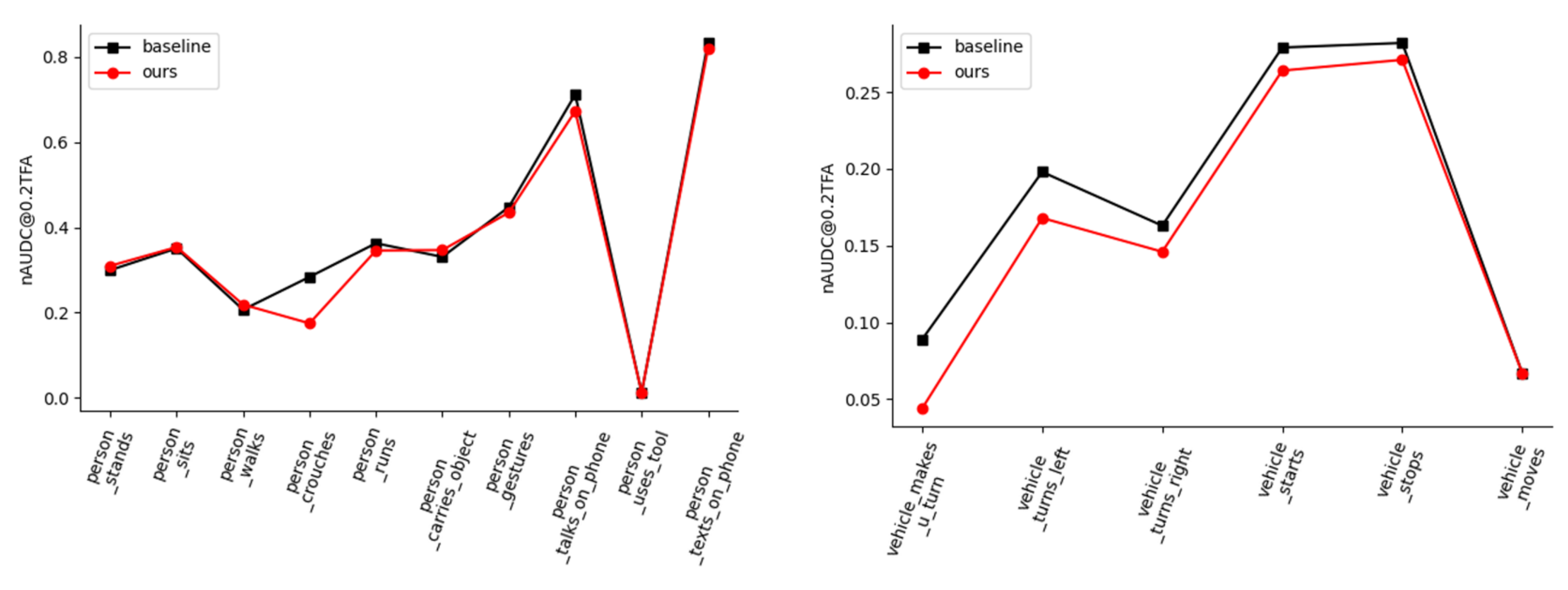}
  \caption{
    The line chart of ablation experiments of our proposed part-attention and localization masking methods.
  }
  \label{figure_ablation}
  \end{figure*}

\section{Method}

\subsection{Overview}

  Figure \ref{figure_framework} (a) shows the overview framework. 
  It consists of four modules: object localizer, proposal filter, activity classifier and activity refiner.
  We give a brief description of them in this section. 
  More details of the activity classifier will be presented in Section \ref{section_person} \& \ref{section_vehicle}.

  \textbf{Object Localizer.}
  Given untrimmed surveillance videos, objects are detected and tracked first. 
  For vehicles, we use Cascade R-CNN \cite{cai2018cascade} to generate bounding boxes (bboxes) per frame. 
  Then DeepSORT \cite{wojke2017simple}, which is a real-time and robust tracking method, is applied to link them into trajectories. 
  In comparison, bboxes of persons tend to be smaller, which is harder to detect if only utilizing frame-level information.
  To exploit inter-frame temporal information fully, we take 3D Cascade R-CNN \cite{li2019effective} as person detector, which outputs clip-level 3D bboxes. 
  Considering bigger bboxes are easier to track, we simply associate these bboxes using IoU like in \cite{bochinski2017high}.
  The trajectories are then used to generate spatio-temporal activities proposals.

  \textbf{Proposal Filter}
  There exist some false detection bboxes in the trajectories. 
  For example, a roadblock maybe detected as a person due to overfitting. 
  Furthermore, static vehicles are generally not related to activities.
  To produce more accurate input proposals for activity classifier, we apply background modeling method to filter out static objects. 
  Specifically, MOG2 \cite{zivkovic2006efficient} is utilized to learn a background model and generate foreground masks. 
  Then a median filter is employed to perform denoising. 
  Finally, proposals with low foreground rates are discarded.

  \textbf{Activity Classifier.}
  We apply two classifiers for person-centered and vehicle-centered individually.
  MViT \cite{fan2021multiscale} pretrained on Kinetics400 \cite{smaira2020short} is applied as the baseline classifier.
  Please see Section \ref{section_person} \& \ref{section_vehicle} for more optimization details.

  \textbf{Activity Refiner.}
  To achieve more accurate localization and classification score, we refine the classified clips inspired by Temporal-NMS \cite{du2021giaotracker}.
  Similar with \cite{yucmu}, the clips are first merged by averaging frame-level confidence scores. 
  Then they are split into spatially and temporally overlapping clips. 
  Finally, those with lower scores are suppressed by using Temporal-NMS, which results in the final outputs.

\subsection{Person-centered Activity Classifier with Part-Attention} \label{section_person}

In surveillance videos, person objects tend to be small and most of person activities are related to their postures and gestures, 
e.g., “person\_crouches”, “person\_talks\_on\_phone” and “person\_gestures”. 
This requires the classifier to be capable of capturing detailed features. 
Moreover, as illustrated in Figure \ref{figure_motivation}, each body part plays a different role for to distinguish different categories.
To this end, we propose a kind of plug-and-play and low-cost part-attention mechanism, as shown in Figure \ref{figure_framework} (b).

Specifically, our part-attention mechanism consists of the following 5 steps:

1) Taking clips $C \in R^{T \times H \times W \times 3}$ as input, we first generate feature maps $F \in R^{T' \times H' \times W' \times C}$ with a feature extractor.
Note that we increase the spatial resolution here by removing its final pooling operation, in order to keep more details.

2) Then a 1x1 convolution layer is applied to output score maps $S \in R^{T' \times H' \times W' \times N}$, where $N$ is the number of classes.

3) The global score $\vec s_1 \in R^{1 \times N}$ is obtained by performing global average pooling (GAP) as usual.

4) Two part scores $\vec s_2, \vec s_3 \in R^{1 \times N}$ are obtained by splitting score maps horizontally and performing global max pooling (GMP). 
The resulting part scores contain more detailed information, and GMP here is essentially a class-specific attention operation \cite{zhu2021residual}.
This could help force the classifier to focus on details. 

5) The final score $\vec s \in R^{1 \times N}$ is the weight sum of $\vec s_1, \vec s_2, \vec s_3$, 
i.e., $\vec s = \vec \lambda_1 \vec s_1 + \vec \lambda_2 \vec s_2 + \vec \lambda_3 \vec s_3$.
The weights $\vec \lambda_i, i=1,2,3$ are vectors instead of scalars, which assign each class a separate weight.
We set $\vec \lambda_1$ as a constant all-one vector and set $\vec \lambda_2, \vec \lambda_3$ as learnable parameters, which are initialized as zero vectors.

The proposed part-attention mechanism could extract more detailed features and learn the contribution of each part features to each class.

\subsection{Vehicle-centered Activity Classifier with Motion Information} \label{section_vehicle}
 Different from person-centered activities, vehicle-centered activities depend more on motion information, e.g., movement direction and speed.
 One common method to encode motion information is optical flow, as in two-stream model \cite{simonyan2014two}. 
 However, it's computationally expensive and sensitive to noise. 
 Another intuitive way is to encode position coordinates in sequence directly using several fully connected layers. 
 But it discards visual information and is hard to optimize jointly with the RGB stream.

 As a consequence, we propose a novel and simple motion information encoding method, as shown in Figure \ref{figure_framework} (c).
 Given an RGB clip $C_1 \in R^{T \times H \times W \times 3}$ and corresponding bounding boxes $B = \{ (x_i, y_i, h_i, w_i) \}_{i=1}^T$,
 we calculate the inter-frame displacement first:
 $$
  (dx_i, dy_i) = 
  \left\{ 
   \begin{aligned}
    & (x_{i+1} - x_i, y_{i+1} - y_i), & if \ i < T \\
    & (0, 0), & if \ i = T
   \end{aligned}
  \right. 
 $$
 Then we initialize an all-zero clip $C_2 \in R^{T \times H \times W \times 2}$ and assign the value $(dx_i, dy_i)$ to the region determined by coordinates $(x_i, y_i, h_i, w_i)$.
 We term clip $C_2$ Motion clip. Finally, we concatenate RGB clip and Motion clip as the input of activity classifier.

 Despite its simplicity, this localization masking method has the following advantages:

 1) Compared with optical flow, our Motion clip is more computationally lightweight. Furthermore, 
 Motion clip doesn't depend on pixel information and only contains two values, i.e., 0 and $dx_i$ (or $dy_i$).
 Thus it is more robust to noise and its motion  information is more consistent.

 2) Different from two-stream model which deals with RGB and motion information separately, 
 we concatenate RGB clip and Motion clip and use a single stream to extract their features.
 It could be thought as a kind of attention mechanism, in which the Motion clip tells the region of foreground objects.
 In other words, it jointly encodes motion and foreground attention information.

 In addition, we argue that the information on the orientation change of vehicles is equally important for activities 
 like “vehicle\_turns\_left” and “vehicle\_turns\_right”. 
 To this end, we perform data augmentation for these two classes by flipping horizontally. 
 Experiments show that our proposed methods could improve the performance significantly.

\begin{table}[t]
  \begin{center}
  \resizebox{0.45\textwidth}{!}{
  \begin{tabular}{cc|c|c|c}
  \toprule[1pt]
      & \textbf{Method} & \textbf{nAUDC(↓)} & \textbf{Person-Centered} & \textbf{Vechicle-Centered} \\ \hline
      & \textbf{baseline} & 0.308 & 0.384 & 0.180  \\ \hline
      & \textbf{ours} & \textbf{0.291} & \textbf{0.369} & \textbf{0.160} \\
  \bottomrule[1pt]
  \end{tabular}}
  \end{center}
  \caption{
    The ablation study of our proposed optimization methods for person-centered and vehicle-centered activities on the VIRAT validation dataset.
  }
  \label{table_ablation}
\end{table}

\begin{table}[t]
  \begin{center}
  \resizebox{0.45\textwidth}{!}{
  \begin{tabular}{cc|c|c|c}
  \toprule[1pt]
      & \textbf{Team} & \textbf{nAUDC} & \textbf{nAUDC@person-centered} & \textbf{nAUDC@vehicle-centered} \\ \hline
      & \textbf{ours} & \textcolor{red}{\textbf{0.409}} & \textcolor{red}{\textbf{0.404}} & \textcolor{red}{\textbf{0.295}} \\ \hline
      & UCF & \textcolor[RGB]{51,204,51}{\textbf{0.431}} & \textcolor[RGB]{51,204,51}{\textbf{0.428}} & \textcolor{blue}{\textbf{0.415}} \\ \hline
      & INF & \textcolor{blue}{\textbf{0.444}} & \textcolor{blue}{\textbf{0.450}} & \textcolor[RGB]{51,204,51}{\textbf{0.391}} \\ \hline
      & M4D & 0.847 & 0.880 & 0.798 \\ \hline
      & TTA & 0.852 & 0.833 & 0.822 \\ \hline
      & UEC & 0.964 & 0.986 & 0.990 \\
  \bottomrule[1pt]
  \end{tabular}}
  \end{center}
  \caption{Leaderboard of TRECVID 2021 ActEV challenge.\textsuperscript{\ref{trecvid21}}}
  \label{table_sota}
\end{table}

\section{Experiments}

In this section, we firstly introduce the VIRAT dataset and the evaluation metric. 
Then we present the ablation study on the VIRAT validation dataset.
Finally, we compare our system to SOTA methods on the VIRAT test dataset.

\subsection{Dataset and Metric}
  The VIRAT dataset \cite{oh2011large} is designed for the task of activity detection (AD) in surveillance videos. 
  Given a target activity, we need to detect and temporally localizes all instances of the activity. 
  Compared with existing action recognition/detection datasets like Kinetics \cite{smaira2020short} and AVA \cite{gu2018ava}, 
  VIRAT brings more challenges in terms of small size of objects, background clutter, diversity in scenes, and complex activity categories. 
  It consists of 64 videos for training, 54 for validation and 246 for testing. 
  35 activity categories are evaluated in the VIRAT dataset, which can be divided into 3 groups, 
  i.e., person-centered, vehicle-centered, interaction activities.
  In this paper, we mainly focus on the 10 categories of person-centered activities and 6 categories of vehicle-centered activities.

  An activity detection system is evaluated by the metric nAUDC@TFA0.2 (for short, nAUDC).
  The system needs to give a confidence score for each activity detected. 
  Then these detected activities are filtered based on this confidence score by a threshold. 
  Varying the threshold makes a trade-off between "being sensitive enough to identify true activity instances (low threshold)" 
  vs. "not making false alarms when no activity is present (high threshold)". 
  This creates a detect error tradeoff (DET) curve for Pmiss (Probability of Missed Detection) and TFA (Time-based False Alarm). 
  The metric nAUDC is calculated as the area under the DET curve across the TFA range between 0\% to 20\% divided by 0.2 to normalize the value to [0,1]. 
  Lower is better as that means fewer errors.

\subsection{Ablation Study}
  Table \ref{table_ablation} and Figure \ref{figure_ablation} present the ablation study results 
  of our person-centered activity classifier and vehicle-centered activity classifier.
  It's shown that our part-attention mechanism exceeds baseline by a large margin, 
  especially for “person\_crouches”, “person\_talks\_on\_phone” and “person\_runs” which depend on detailed features. 
  That demonstrates our part-attention could strengthen the ability of the classifier to focus on details.
  As for vehicle-centered activities, the proposed localization masking method benefits all 5 vehicle-only activities 
  (except for “vehicle\_moves” which only relys on the quality of proposals).

\subsection{Comparison to SOTA}
  Moreover, table \ref{table_sota} illustrates the leaderboard of the TRECVID 2021 ActEV challenge and our team wins the 1st place.
  \footnote{https://actev.nist.gov/trecvid21\#tab\_leaderboard\label{trecvid21}}
  Specifically, our method achieves the best performance for both person-centered and vehicle-centered activities.

\section{Conclusion}

In this paper, we introduce an activity detector system for person-centered and vehicle-centered activities in surveillance videos, which consists of four modules. 
Particularly, we propose a novel part-attention mechanism which forces the person-centered activity classifier to focus on detailed features of different body parts. 
As for vehicle-centered activities, a simple but effective motion information encoding method is introduced. 
Our system achieves the best performance on the VIRAT dataset and our team wins the 1st place in the TRECVID 2021 ActEV challenge.

\section{Acknowledgements}
   This work is supported by Chinese National Natural Science Foundation under Grants (62076033, U1931202).

\bibliographystyle{IEEEbib}
\bibliography{icme2021template}
\end{document}